\documentclass[final]{cvpr}

\usepackage{times}
\usepackage{epsfig}
\usepackage{graphicx}
\usepackage{amsmath}
\usepackage{amssymb}
\usepackage{booktabs}

\usepackage[normalem]{ulem}

\usepackage[pagebackref=true,breaklinks=true,colorlinks,bookmarks=false]{hyperref}

\def\cvprPaperID{19} 
\def\confYear{CVPR 2021}

\begin{document}

\title{eProduct: A Million-Scale Visual Search Benchmark to Address Product Recognition Challenges}

\author{Jiangbo Yuan, An-Ti Chiang, Wen Tang, Antonio Haro\\
eBay Inc., San Jose, CA, USA\\
{\tt\small \{jiayuan, anchiang, wtang4, anharo\}@ebay.com}
}

\maketitle

\begin{abstract}
Large-scale product recognition is one of the major applications of computer vision and machine learning in the e-commerce domain. Since the number of products is typically much larger than the number of categories of products, image-based product recognition is often cast as a visual search rather than a classification problem. It is also one of the instances of super fine-grained recognition, where there are many products with slight or subtle visual differences. It has always been a challenge to create a benchmark dataset for training and evaluation on various visual search solutions in a real-world setting. This motivated creation of eProduct, a dataset consisting of 2.5 million product images towards accelerating development in the areas of self-supervised learning, weakly-supervised learning, and multimodal learning, for fine-grained recognition. We present eProduct as a training set and an evaluation set, where the training set contains 1.3M+ listing images with titles and hierarchical category labels, for model development, and the evaluation set includes 10,000 query and 1.1 million index images for visual search evaluation. We will present eProduct's construction steps, provide analysis about its diversity and cover the performance of baseline models trained on it.
\end{abstract}
\vspace{0.1cm}

\section{Introduction}
Given the prevalence of online shopping in daily life, it is imperative for computer vision systems to automatically and accurately recognize products based on images. The primary objectives of product recognition include building a catalog of products, similar product recommendations, and visual or text product search, to name a few. These improve product discoverability, increase seller and buyer engagement and conversions for e-commerce \cite{ecommerce-search-2020}.

\begin{figure}[tb]
\includegraphics[width=1.0\linewidth]{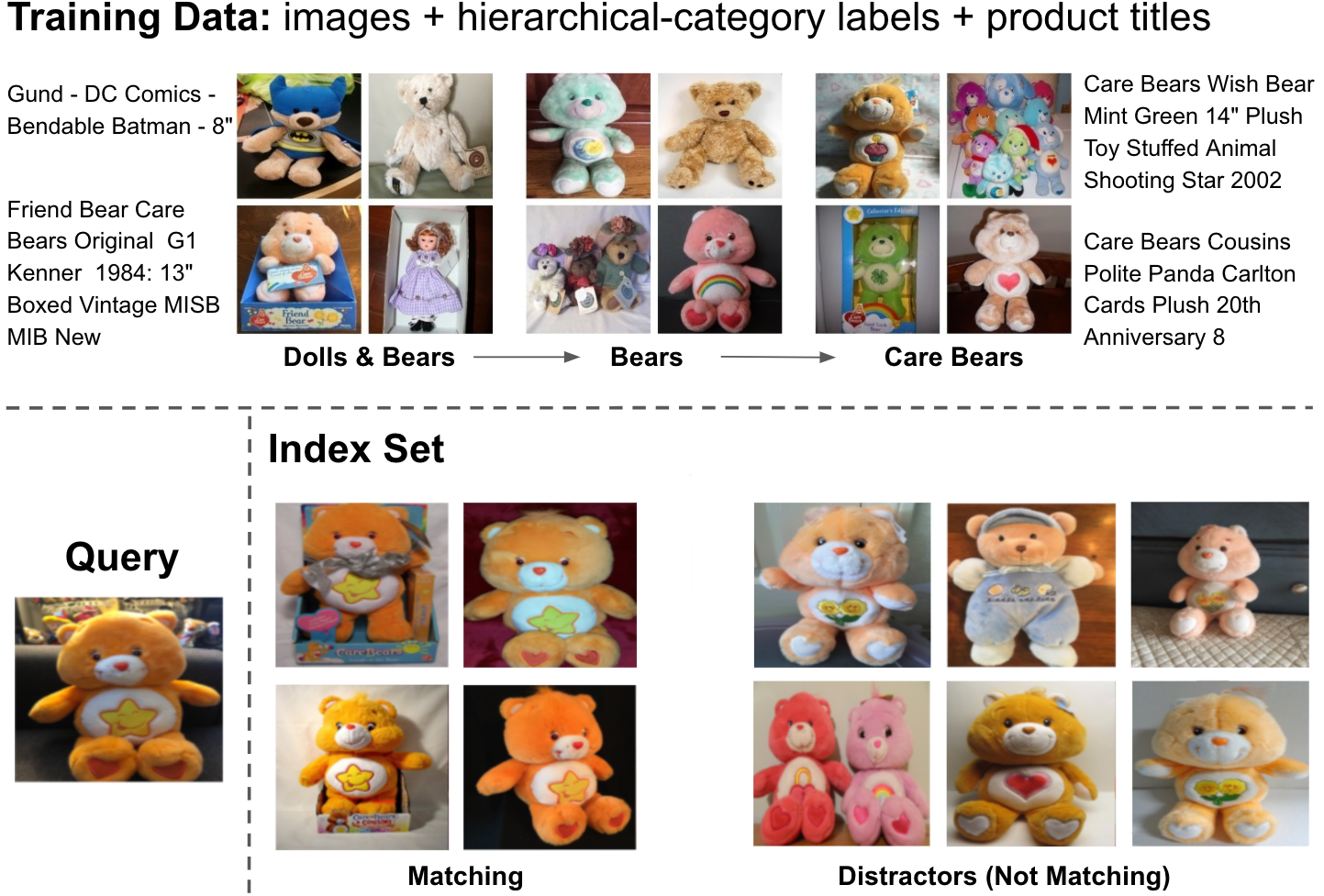}
\caption{eProduct structure illustration: Instead of expensive product-level labels, eProduct provides a hierarchy of weakly-labeled categorical annotations and image titles for model training. \textbf{Top:} Example showing the label hierarchy of eProduct via a meta-to-leaf branch from the “Dolls \& Bears" subtree. For each synset, 4 randomly sampled images are presented. Title examples are given for the boundary images. \textbf{Bottom:} Example of eProduct's visual search evaluation data: we show a query image on the left with its four true matches and six distractors on the right. Distractors are “hard" examples because they all come from the same leaf category as the query, i.e. “Care Bears, Teddy Bears", yet only the true matches share the same product model--“Care Bears Laugh-a-lot Bear Orange Star"--with the query. The dataset exposes a gap between the extremely fine-grained recognition and the weakly-labeled training data, which challenges many current existing learning mechanisms.}
\label{fig:eproduct-illustration}
\end{figure}

Product recognition at scale is a major challenge for buying or selling consumer goods on online service platforms. The first challenge is to create a large-scale, diverse, well-distributed and cleanly annotated dataset for training given a large number of likely fine-grained categories, many with subtle differences not easily distinguishable. Models trained on available generic image datasets \cite{labelme-tool, lin2015microsoft, imagenet, 80m-tiny-images, OpenImage-v4} are often sub-optimal and not desirable in real-applications. Although there is some released e-retail data, it is only specialized on fashion \cite{liu2016deepfashion, deepfashion2, WhereToBuyItICCV15, FashionMNIST}. It becomes more difficult when thousands of new products are added to the inventory every day, which also leads to a quick degradation of the off-line trained model. This is an instance of open-world learning. Like in most practical customer to customer marketplace use-cases, we assume that coarse category labels or text descriptions (e.g. image titles) are easily available while the super-fine product labels are not. Moreover, a fair model evaluation is also critical in the machine learning development life-cycle. However, to date, there is still a need for a visual search benchmark for product recognition with considerations on label accuracy, object diversity and scale.


To that end, \textit{eProduct}, a large-scale product image dataset was created to accelerate advances in super fine-grained product recognition. We prioritize the following goals: 1) To support a variety of learning paradigms, we provide both text (product titles) and categorical labels alongside the images; 2) To address the need for large-scale product recognition benchmarking, we constructed a visual search benchmark consisting of 10k query images (covering 206 leaf categories), 100k groundtruth match images, and another 1.1 million distractor images for exact product matching evaluation; 3) To maximize reuse of existing machine learning algorithms with this new dataset, we constructed training sets inspired by the ImageNet-1M dataset.

More specifically, eProduct uses the hierarchical taxonomy structure which is a sub-tree from a real and large e-commerce inventory. It includes 16 meta categories and 1,000 leaf categories, and 1.3 million images associated with their titles for training use, 10,000 query images alongside 1.1 million index images are used for a visual search benchmark. Its statistics are shown in Table \ref{tab:eproduct-statistics}. Images were collected with different qualities ranging from user uploaded photos to professional settings. Groundtruth matches in the visual search benchmark often come from multiple sellers or listings, which reduces bias when evaluating retrieval models. The setting of eProduct may be broadly interesting to many computer vision research areas, such as, multi-modal learning based on visual and text information \cite{CLIP, DallE2021}, weakly-supervised learning \cite{weakly-supervised-review}, self-supervised learning on images \cite{SimCLR-v1, SimCLR-v2, MOCO-v1, MOCO-v2, SwAV2021}. Moreover, eProduct also provides a technical benchmark to fairly evaluate the progress of the latest computer vision techniques. Fig. \ref{fig:eproduct-illustration} illustrates the eProduct structure.

\begin{table}[tb]
\small
\centering
\begin{tabular}{c|ccccc}
\toprule
Subset & \# of Images & Titles & Meta & Level-2 & Leaf \\ 
\midrule
Train   & 1,281,167 & Yes & 16  & 75  & 1000       \\ 
Val    & 50,000  & Yes  & 16  & 75  & 1000      \\ 
\midrule
Query   & 10,000  & -   & -  & -  & -     \\ 
Index   & 1,101,396 & -  & -  & -  & -     \\ 
\bottomrule
\end{tabular}
\caption{eProduct statistics.}
\label{tab:eproduct-statistics}
\end{table}
\vspace{0.1cm}

\section{Training Data Construction}
eProduct provides \textit{train} and \textit{val} subsets with hierarchical category labels and image titles. In the context of visual search, the two subsets together will serve as the training data for model development. It is worth noting that no product-level labels are provided in the training but the visual search is at product level. The gap significantly challenges current solutions yet provides opportunities for novel weakly-supervised learning, multimodal learning and others.

\begin{figure}[tb]
 \includegraphics[trim={0cm 0 0 0.5cm},width=0.95\linewidth]{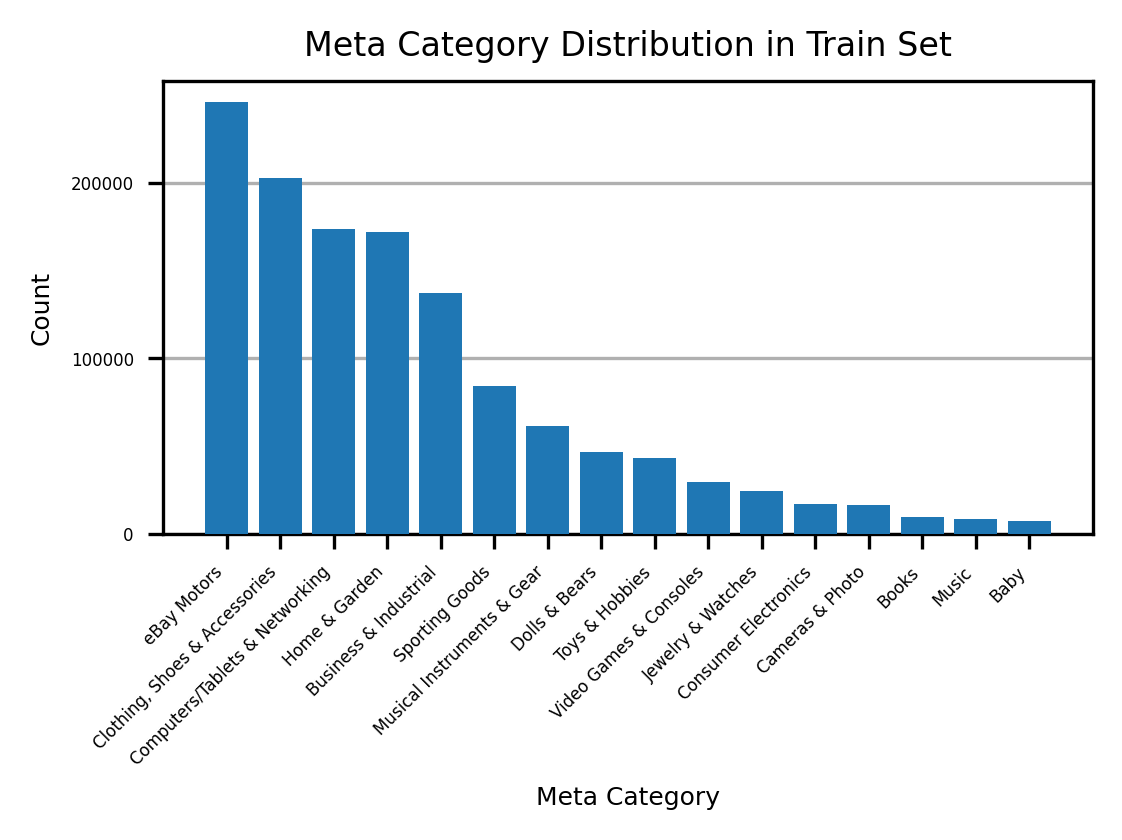}
 \caption{Training data distribution over meta category.}
 \label{fig:training-set-meta-distribution}
\end{figure}

\begin{figure}[tb]
 \includegraphics[trim={0cm 0 0 0.5cm},width=0.95\linewidth]{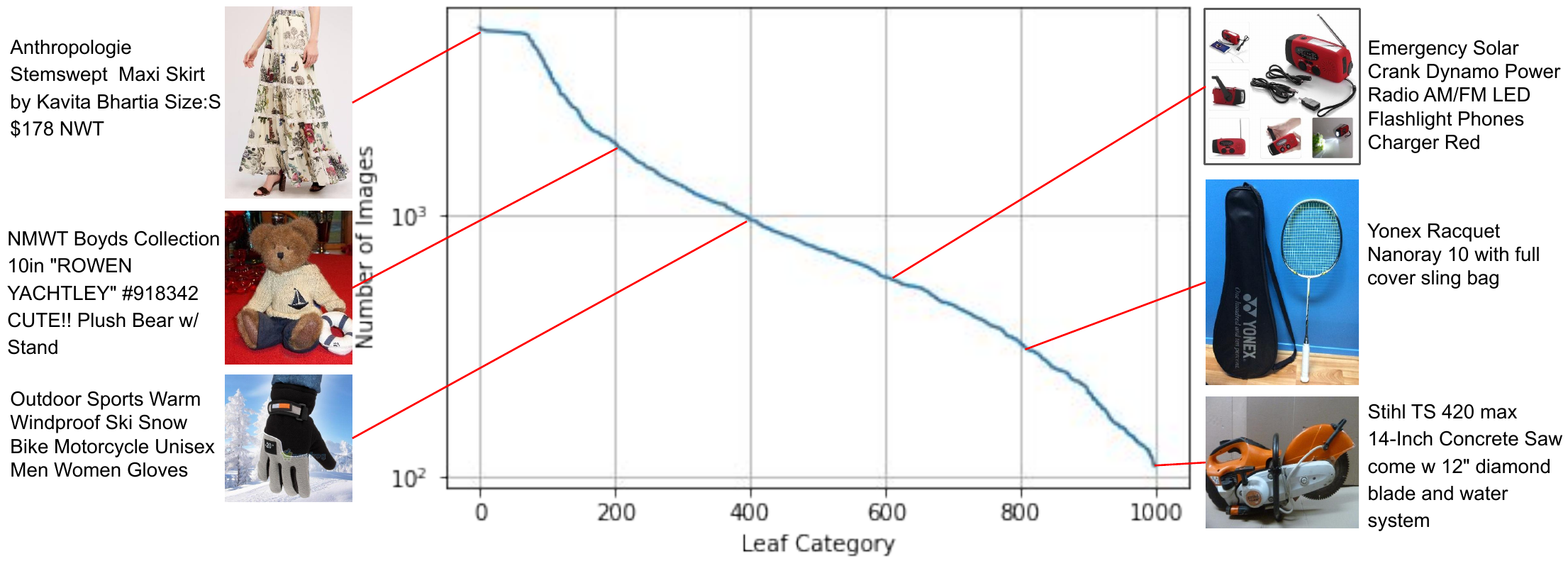}
 \caption{Training data distribution over leaf category. eProduct contains a large imbalance between classes. We clipped the largest classes to rebalance the overall distribution. The plateau on left reflects the result of the clipping process for data rebalancing.}
 \label{fig:training-set-examples}
\end{figure}

To construct the training set, we randomly selected 1000 leaf categories from 16 meta categories from eBay3500, an internal 25 million image dataset which covers 27 meta and 3,500+ leaf categories. Similar to many other fine-grained datasets, a long-tailed distribution is common in such datasets. However, we have clipped the top of the largest leaf categories to reduce their dominance. More specifically, the sampling with clipping strategy was performed in three steps: 1) 1,000 leaf categories were randomly sampled from eBay3500 and resulted in about 10 million images; 2) leaf categories were each clipped by a maximal threshold; 3) train/val subsets, with the same sizes as ImageNet-1M, were created by randomly sampling from the output of step 2. The sampled 16 meta-category distribution of the train set can be found in Fig. \ref{fig:training-set-meta-distribution}. Due to the clipping process, the data distribution is rebalanced by restricting the super-large categories and results in raising super-tiny ones. The rebalancing is more motivated from real practices, where we try to create less imbalanced training data for better model performance. Fig. \ref{fig:training-set-examples} shows the training data subset distribution over leaf categories along with sample images from different class scales.

\section{Visual Search Benchmark Construction}\label{visual-search-construction}

To construct a good evaluation benchmark, three important factors are taken into account: 1) \textbf{Query Coverage:} not only the training categories but also the out-of-sample categories are included in the query set to enhance the product diversity and searching difficulties. 2) \textbf{High Quality of Matches:} each query has on average 10.2 exact matches coming from multi-sellers, and the images differ in angles, backgrounds and clutter, object size, occlusions, and illumination; 3) \textbf{Difficulty:} a large amount of distractors is added as noise to make the dataset challenging.

\subsection{Query Set}
\begin{figure}[tb]
\centering
 \includegraphics[trim={0cm 0 0 0.5cm},width=\linewidth]{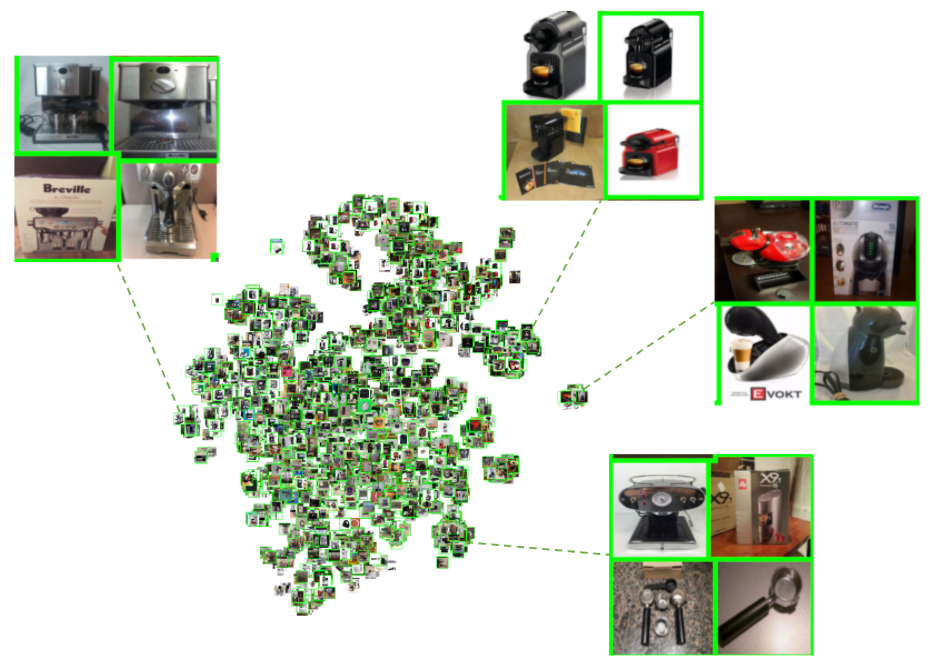}
 \caption{Query sampling from title embedding space with consideration of diversity. Green rectangles represent the picked query samples.}
 \label{fig:query_sampling}
\end{figure}

Diversity was prioritized when collecting the queries. We thus manually selected 500 leaf categories inspired by an in-house catalog database, where over 200 leaf categories are found in eBay3500, 16 meta and 156 leaf categories overlap with the training data. In order to sample efficiently from such a large dataset, we first constructed title embeddings for all candidates and then sampled in each leaf category following a method similar to $k$-means++ \cite{kmeansplusplus} seeding strategy, as shown in Fig. \ref{fig:query_sampling}, aimed to avoid ubiquitous product queries. For each title embedding, we used average pooling on all of the corresponding fastText \cite{joulin2016fasttext} word embeddings. The word embedding was processed in the following manner: 1) fastText vocabulary size is 600K; 2) stop words were removed; 3) PCA was then applied which resulted in final 64-D vectors reduced from 300-D. 

\subsection{Index Set}

The index set is the search space in our visual search task. It implicitly consists of two subsets, \textit{groundtruth matches} and \textit{distractors}. A groundtruth match is an exact product match to any of the queries, while a distractor image is not a match to any queries. A clean and large-scale index set makes a visual search dataset valuable in terms of accurate evaluations and representative predictions for real-world applications. 

In order to collect both matches and distractors, we first created a candidate pool by randomly sampling from eBay3500 following a similar process as the Train set. Next we need to annotate all candidates to be either “exact match" or “not-a-match". The labeling cost for a complete human annotation is prohibitively expensive. To minimize the cost, we used a preranking methodology to reduce the size of shortlists for human review. The preranking candidates were processed in several steps including categorical filtering, similarity-based thresholding, and a newly proposed multi-modal asynchronous-voting method to further improve the recalls. With all these processes, we created candidate shortlists of various lengths for final human-supervised verification.

The final average size for the shortlists was about 200 candidates, 0.02\% of the original size. During the crowd-sourced annotation process, given query-candidate pairs, human annotators reviewed both image and title labeling at several levels of confidence for “exact match”. Each query-candidate pair has been rated by at most ten annotators. We performed post processing to determine the final acceptable labels. We also discarded any labels with low confidence as either “match” or “not a match” to minimize incorrect annotation risk. We also collected an initial match set via different methodologies including manually querying from eBay listings. This initial set has been merged into the above candidate pool before further processing.
\\

\noindent\textbf{Human Annotation} The goal of the human annotation is to identify exact matches (same product) and non-matches based on the given preranking results. However, the labeling decisions are not a binary problem. Even though we have both images and titles available for human annotators, in many cases, they are not able to decide a label if we only enable binary options. For that reason, we added a third option, the “Maybe" label. We next introduce our definitions of \textit{same product} to aid understanding of the groundtruth label construction.

\noindent\textbf{Same Product} 
To define two images as including the “same” product is often subjective and challenging especially when critical aspects are missing from their titles and are invisible from images. In our case, we use images as the primary source for annotations. This enables fairer model performance evaluation for visual search. Titles have been also used to establish products being “same” in many cases. The criteria for “same product" in two product images is:
\begin{itemize}
    \item Their two images can be different (different angles, different backgrounds etc.) but the products in these images should be the same.
    \item The two products can represent different product conditions. For example, a broken phone and a new phone with the exact same specifications (make, model, color, etc.) are the “same products”.
    \item They should have the same color/model/style, and other aspects that are visually visible and distinguishable. For example, a golden and a gray phone of otherwise the same make and model are not considered the “same products”.
    \item They can vary in other aspects that are not distinguished solely based on images, e.g., shoe size, memory size for hardware, etc.
\end{itemize}

\begin{figure}[tb]
\centering
 \includegraphics[width=0.95\linewidth]{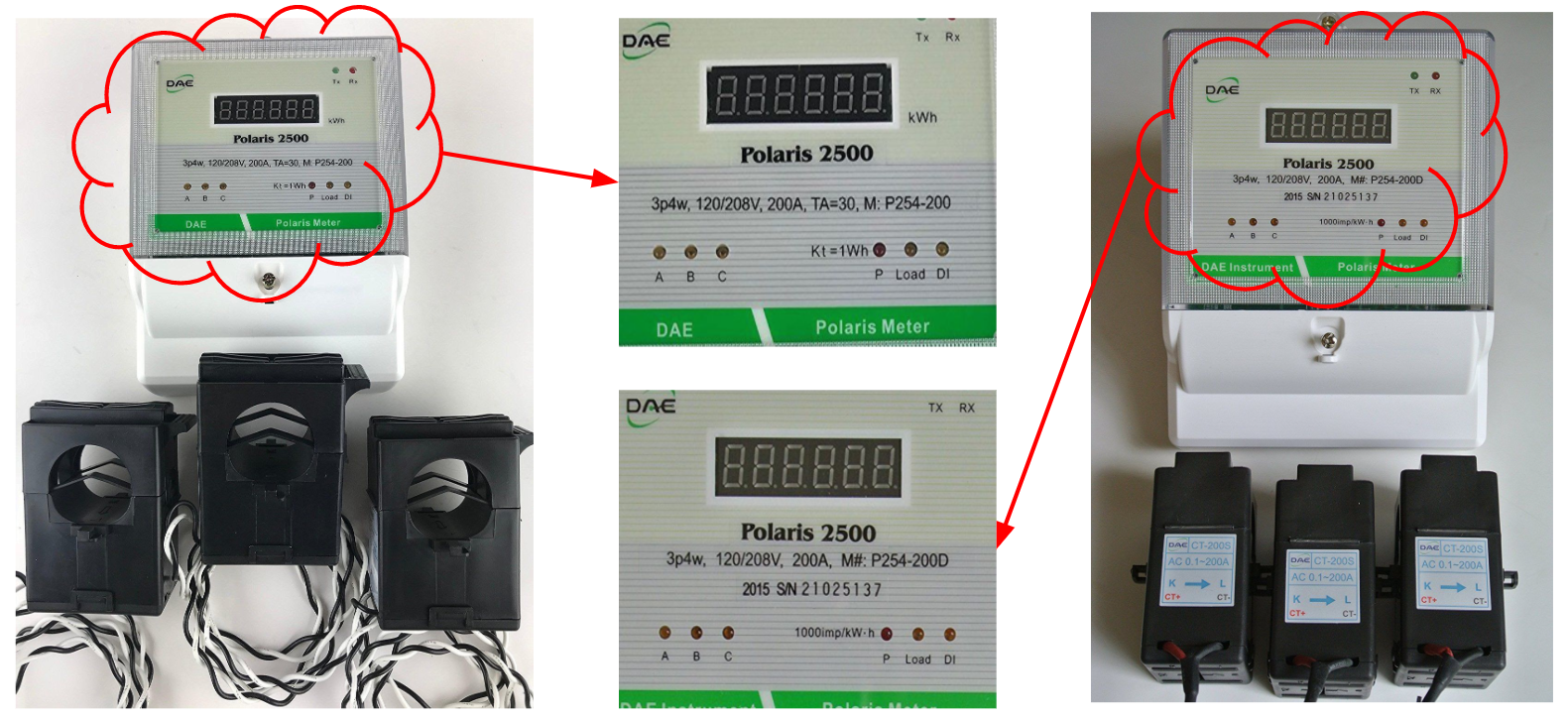}
 \caption{A hard example which was labeled as “exact match" in our initial match set. It was corrected in the later stage after our preranking and crowd-sourced annotation process.}
 \label{fig:labeling-hard-example}
 \vspace{0cm}
\end{figure}

Differing from the single worker annotation for the initial match set, we had up to ten annotators from a commercial crowd-sourcing platform  annotate each pair for possible matching. Final label decisions were made based on a voting method with statistically selected thresholds, while jointly utilizing a small scale but carefully selected "Golden Dataset" and spot-checks. All samples, with either “Exact Match" or “Not a Match", were merged into the rest of not-a-match candidates after the first stage preranking. The combination set forms our eProduct Index set. Note that, tens of thousands of samples with “Maybe" labels were discarded from the final eProduct dataset. About 40\% of the initial match set has been removed from final groundtruth matches, showing how difficult it is to have reliable and accurate labels from a single annotator. Fig. \ref{fig:labeling-hard-example} shows such an example.


\section{Evaluation Criteria for Visual Search}
We here focus on the evaluation of visual search tasks. Given a query image, the goal is to retrieve the images from the million-scale Index set, to produce rankings that list as many of the possible exact matches on the top. 

Our major criterion is Macro-Average Recall@$k$, or $\mathbf{MAR}@k$, and we often choose $k=10$. $\mathbf{MAR}@k$ is simply the average of recall@k over all queries, and the higher the better. The mathematical definition is:
\vspace{0.05cm}
$$\mathbf{MAR}@k = \frac{1}{N} \sum_{i=1}^{N} \mathbf{R}_{i},$$
\vspace{0.05cm}
where $N$ is the total number of queries. $\mathbf{R}_{i}$ represents $recall@k$. Given a query \textit{i}, $recall@k$ is the proportion of true match items found in the top-$k$ recommendations. Mathematically, $recall@k$ for the $i$-th query is defined as follows: $\mathbf{R}_{i} = \frac{r_{i}}{g_{i}}$, where: $r_{i}$ is the \# of recommended items at $k$ that are true matches to query $i$. 

Note that the number of groundtruth matches for a given query is at least 1 and could be greater than $k$. Thus $g_{i} \in [1, k]$, i.e., min($k$, \# of groundtruth matches for query i). And a correct recommendation or a true match is an item from the top-$k$ result that matched any items, not only the top-$k$, from the full groundtruth list. This metric and evaluation setting is similar to a real production scenario where a product might have many more matches but we are usually focused on the recalls at the top-$k$'s.\\

\noindent \textbf{ResNet Baselines}: To validate the evaluation metric, we trained ResNet-50 and ResNet-101 \cite{he2016deep} models on eProduct from scratch. SGD optimizers with momentum of 0.9 and $10^{-4}$ weight decay were used for training over 95 epochs. The first 5 epochs were additionally used to warm-up the learning by linearly increasing the learning rates. The learning rate was respectively scaled down by 0.1 decay at the 30-th, 60-th and 80-th epochs. Global average features were extracted and tested on the eProduct visual search task, which achieved $\mathbf{MAR@10=0.4091}$ and $\mathbf{MAR@10=0.4540}$ for ResNet-50 and ResNet-101, respectively.

\section{Conclusion}
A diverse e-commerce inventory consists of tens of thousands of categories and millions or billions of different products. However, the attention from published benchmarks has been focused on small scale subsets or specific categories. eProduct creates opportunities for novel research, with shared robust training and evaluation datasets, to enable advanced computer vision and machine learning models with results predictive of real-world product recognition systems.

{\small
\bibliographystyle{ieee_fullname}
\bibliography{egbib}
}

\end{document}